\documentclass[10pt,final,3p]{elsarticle}
\usepackage{lmodern}

\usepackage{graphics}
\usepackage{graphicx}

\usepackage{booktabs}
\usepackage{multirow}
\usepackage{array}
\usepackage{amsmath}

\usepackage{amsthm}
\usepackage{verbatim}
\usepackage{relsize}
\usepackage[dvipsnames]{xcolor}
\usepackage{array}
\usepackage{bm}
\usepackage{caption}

\usepackage{algorithm}
\usepackage{algpseudocode}
\usepackage{enumitem}

\usepackage{natbib}
\usepackage{hyperref}
\newcommand{\doi}[1]{\textsc{doi}: \href{http://dx.doi.org/#1}{\nolinkurl{#1}}}

\definecolor{comment}{rgb}{0.0, 0.5, 0.0}

\usepackage{hyperref}

\hypersetup{
    colorlinks=true,
    linkcolor=blue,
    filecolor=magenta,
    urlcolor=cyan,
}
\usepackage{booktabs}
\usepackage{xcolor}
\usepackage{color}

\definecolor{mygreen}{rgb}{0,0.6,0}
\definecolor{mygray}{rgb}{0.5,0.5,0.5}
\definecolor{mymauve}{rgb}{0.58,0,0.82}

\usepackage{pdfcomment}

\usepackage{array}
\usepackage{mathtools}
\usepackage{wasysym}
\usepackage{verbatim}
\usepackage{relsize}
\usepackage{multirow}
\usepackage{multicol}

\usepackage{listings}
\usepackage[bitstream-charter]{mathdesign}
\usepackage[T1]{fontenc}
\usepackage[utf8]{inputenc} 

\usepackage[english]{babel}
\usepackage[normalem]{ulem}

\DeclareMathAlphabet{\pazocal}{OMS}{zplm}{m}{n}

\def \A {\pazocal{A}}

\usepackage{lscape}		
\usepackage{fancyvrb}

\usepackage[mathscr]{euscript}
\usepackage{textcomp}

\def \balpha {{\pmb{\alpha}}}
\def \bbeta  {{\pmb{\beta}}}
\def \bzero  {{\pmb{0}}}

\def \bPsi {{\boldsymbol\Psi}}
\def \xxi  {{\boldsymbol{\xi}}}
\def \X  {\pmb{\pazocal{X}}} 

\newcommand{\Ns}{{\ensuremath{n_{\mathrm{sim}}}}}

\def \PCE {\textsf{PCE}}
\def \PC {\textsf{PC}$^2$}

\usepackage{bigints}

\usepackage[switch]{lineno}

\usepackage{lineno}

\begin{document}

\begin{frontmatter}



\title{Physics-Informed Polynomial Chaos Expansions}


\author{Luk{\'a}{\v s} Nov{\'a}k\corref{cor1}}  \ead{novak.l@fce.vutbr.cz}  \cortext[cor1]{Corresponding author}
\address{Brno University of Technology, Brno, Czech Republic}
\author{Himanshu Sharma, Michael D. Shields}                     \ead{\{hsharm10, michael.shields\}@jhu.edu}
\address{Johns Hopkins University, Baltimore, USA}

\begin{abstract}
Surrogate modeling of costly mathematical models representing physical systems is challenging since it is typically not possible to create a large experimental design. Thus, it is beneficial to constrain the approximation to adhere to the known physics of the model. This paper presents a novel methodology for the construction of physics-informed polynomial chaos expansions (\PCE{}) that combines the conventional experimental design with additional constraints from the physics of the model. Physical constraints investigated in this paper are represented by a set of differential equations and specified boundary conditions. A computationally efficient means for construction of physically constrained \PCE{} is proposed and compared to standard sparse \PCE{}. It is shown that the proposed algorithms lead to superior accuracy of the approximation and does not add significant computational burden. Although the main purpose of the proposed method lies in combining data and physical constraints, we show that physically constrained \PCE{}s can be constructed from differential equations and boundary conditions alone without requiring evaluations of the original model. We further show that the constrained \PCE{}s can be easily applied for uncertainty quantification through analytical post-processing of a reduced \PCE{} filtering out the influence of all deterministic space-time variables. Several deterministic examples of increasing complexity are provided and the proposed method is applied for uncertainty quantification.
\end{abstract}




\end{frontmatter}


\section{Introduction}
\label{}

Mathematical models of real-life physical systems are typically highly computationally demanding and contain various uncertain variables, thus it is necessary to use surrogate models as computationally cheap approximations to perform \emph{uncertainty quantification} (UQ), optimization, parametric studies, and other tasks. These surrogate models treat the original model as a black-box and fit the model from several deterministic simulations at given data points in the design domain. Their practical use requires a sufficient number of data points covering the space of input random variables, which can be prohibitively expensive. However, it is beneficial to incorporate additional constraints, e.g., \ known physical principles, to ensure realistic and physically meaningful surrogate model behavior and reduce training data demands. 

Recently, there has been a considerable interest in developing machine learned surrogate models capable of satisfying physical constraints -- spawning an entirely new field of \textit{physics-informed machine learning} \cite{karniadakis2021physics}. Much of this work has focused on neural network models, including the hugely popular physics-informed neural networks (PINNs) \cite{raissi2019physics} and the related physics-informed neural operators \cite{goswami2022physics}. Numerous variants of deep neural networks with physical constraints are now available, including physics-informed autoencoders \cite{erichson2019physics}, the Deep Ritz Method \cite{kharazmi2019variational}, and Deep Galerkin Method \cite{sirignano2018dgm} to name just a few.  In fact, the recent developments and applications of PINNs and related neural networks are so numerous we cannot possibly cover them here. But neural networks are not the only machine learning models that benefit from physical constraints. Recent studies have also incorporated physical constraints into Gaussian process regression models \cite{swiler2020survey, pang2020physics}, which have the benefit of providing a natural measure of uncertainty. These physically constrained Gaussian processes have been applied for UQ of highly complex physical systems~\cite{sharma2023learning}.

This paper focuses on imposing constraints on Polynomial Chaos Expansions (\PCE) \cite{Wiener_PCE}, which we treat here as a machine learning regression method \cite{LuthenReview, torre2019data}. Our interest in \PCE, stems from its usefulness for UQ tasks that include moment estimation and sensitivity index computation \cite{SUDRET}, as well as its convenient properties for surrogate modeling in UQ that derive from their orthogonality properties with respect to certain probability measures of input variables. Although \PCE{} surrogate models have distinct benefits for UQ \cite{CRESTAUX20091161,CHEN20084830}, they suffer from the \emph{curse of dimensionality} since the number of \PCE{} terms grows rapidly with both dimension and maximum polynomial order. Naturally, this significantly affects the necessary number of data points needed for regression-based \PCE{}. As a result, recent research has focused on exploiting the analyzed mathematical model to minimize the size of the basis set and maximize the information obtained from each data point.  This involves the construction of optimal sparse basis by adaptive algorithms \cite{Lüthen_2022}, determining the optimal positions of data points for regression \cite{PCESoptimSeq,NOVAK2021114105,SequentialPCEThapa} and constructing various types of localized surrogates consisting of many low-order \PCE s  \cite{WAN2005617,novák2023active,SSE:Marelli:21}. In addition to the information obtained directly from the mathematical model, there are often specific physical characteristics and constraints derived from the nature of the quantity of interest (QoI) that should be incorporated into the PCE. Therefore we propose a novel and computationally efficient method to incorporate known physics in the form of ordinary differential equations (ODEs) and partial differential equations (PDEs) and their boundary conditions (Dirichlet, Neumann, or mixed) in the \PCE framework, referred to as physically constrained polynomial chaos (\PC{}) expansion. We enforce the constraints at discrete points in the input domain, referred to as virtual points, to solve a constrained least square optimization problem for the \PC{} coefficients. This approach can be generalized to incorporate any form of physical constraints.

Exploiting known physics and boundary conditions in the construction of a \PCE \ approximation assures higher prediction accuracy, especially in regions of input space containing an insufficient number of training points. In the proposed \PC{} framework, we can generally adopt any numerical optimization technique to solve the constrained optimization problem. However, as the number of \PC{} coefficients and virtual points increases, some of the numerical optimization techniques could lead to convergence issues or their computational cost could be significant, which restricts its use to high dimensional problems.
In this paper, we focus specifically on equality-type constraints with the objective of minimizing the sum of squared residuals (SSR), and thus, it is possible to use the well-known method of Lagrange multipliers. The Karush-Kuhn-Tucker (KKT) stationarity condition yields a system of equations that can be solved for the \PC{} coefficient. This approach is a natural extension of the ordinary least squares (OLS) approach, as used in \PCE{} \cite{Choi2004}, to incorporate known equality-type constraints in a computationally efficient manner. We further show that the approach can be combined with sparse regression methods, specifically the Least Angle Regression (LAR)~\cite{BLATMANLARS} to efficiently reduce the basis set. We then apply the method to several benchmark ODEs and PDEs to illustrate its performance. Finally, we demonstrate how it can be used for general UQ purposes.

\section{Non-intrusive Polynomial Chaos Expansion}

\label{PCE section}

Assume a~probability space ($ \Omega, \pazocal{F}, \pazocal{P} $), where $ \Omega $ is an event space, $ \pazocal{F} $ is a~$ \sigma $-algebra on $ \Omega $ and $ \pazocal{P} $ is a~probability measure on $ \pazocal{F} $.
If the input variable of a~mathematical model, $Y=u(\pazocal{X})$, is a~random variable $ \pazocal{X}(\omega) , \omega\in\Omega$,
the model response $Y$($ \omega $) is also a~random variable. Assuming that $ Y $ has finite variance, \PCE\ represents the output variable $ Y $ as a~function of an another random variable $ \xi $ called the \emph{germ} with a~known distribution
\begin{equation}
    Y=u(X)=u^{\PCE}(\xi ),
\end{equation}
and represents the function $u(\pazocal{X})$ by an infinite expansion of polynomials that are orthogonal with respect to the probability density $p_\xi(\xi)$ of the germ. 
These polynomials form a basis of the Hilbert space  $ L^2 $ ($ \Omega,  \pazocal{F}, \pazocal{P} $) of all real-valued random variables of finite variance, where  $ \pazocal{P} $ takes over the meaning of the probability distribution. The orthogonality condition is given by the inner product on $ L^2 $~($ \Omega,  \pazocal{F}, \pazocal{P} $) defined for any two functions $ \psi_j $ and $ \psi_k $ with respect to the weight function $ p_\xi(\xi) $ as:
\begin{equation}
    \langle	\psi_j,\psi_k\rangle
    =
    \int \psi_j(\xi)\psi_k(\xi)p_\xi  (\xi)
    \; \mathrm{d} \xi
    = 0 \quad \forall j\ne k.
\end{equation}

This means that there are specific orthogonal polynomials associated with the corresponding distribution of the germ. Orthogonal polynomials corresponding to common distributions can be chosen according to the Wiener-Askey scheme~\cite{Askey}. For further processing, it is beneficial to use orthonormal polynomials, where the inner product of polynomials is equal to the Kronecker delta  $ \delta_{jk}$, i.e. $ \delta_{jk}=1$ if $ j=k $, and  $ \delta_{jk}=0 $ otherwise.

In the case of $ \X $ and $\xxi$ being vectors containing $ M $ independent random variables, the polynomial basis $ \Psi (\xxi)$ is multivariate and it is built up as a~tensor product of univariate orthonormal polynomials, i.e.
\begin{equation}
\label{Eq: MultVarPol}
    \Psi_{\balpha} ( \xxi )
    =
    \prod_{i=1}^{M}  \psi_{\alpha_i}(\xi_i),
\end{equation}
where $ {\balpha}\in \mathbb{N}^M $ is a~set of integers called the \emph{multi-index} reflecting polynomial degrees associated to each $\xi_i$. The quantity of interest (QoI), i.e. the response of the mathematical model $ Y=u(\X)$, can then be represented as~\cite{Ghanem_spectral}
\begin{equation}
\label{PCE}
    Y = u(\X) =
    \sum_{\balpha \in \mathbb{N}^M }
    \beta_{\balpha}\Psi_{\balpha}( \xxi),
\end{equation}
where $ \beta_{\balpha} $  are deterministic coefficients and $ \Psi_{\balpha} $ are multivariate orthonormal polynomials.

For practical computation, the \PCE\ expressed in Eq.~\eqref{PCE} must be truncated to a~finite number of terms, $ P $. One can generally choose any truncation rule (e.g. tensor product of polynomials up to the selected order $p$), but the most common truncation is achieved by retaining only terms whose total degree $ \vert \balpha \vert $ is less than or  equal to a~given $ p $, in which case the truncated set of \PCE\ terms is then defined as
\begin{equation}
    \pazocal A^{M,p}
    =
    \left\{
        {\balpha} \in \mathbb{N}^{M} : \left| {\balpha} \right|= \sum_{i=1}^{M} \alpha_i \leq p
    \right\}.
\label{Eq: truncation}
\end{equation}
The cardinality of the truncated \emph{index set} $ \pazocal A^{M,p} $ is given by
\begin{equation}
 \mathrm{card} \: \pazocal A^{M,p}= \frac{\left( M+p \right)!}{M! \: p!}\equiv P \, .
 \label{Eq.: Cardinality PCE}
\end{equation}

When the \PCE\ is truncated to a~finite number of terms, there is an error $ \varepsilon $ in the approximation such that

\begin{equation*}
    Y
    =
    \displaystyle u{(\X)} =
    \sum_{\balpha \in \pazocal A}
    \beta_{\balpha} \Psi_{\balpha}(\xxi)
    +
    \varepsilon  \, = Y^{PCE} +\varepsilon.
    \label{Eq: Chaos_Truncated_OLS}
\end{equation*}
From a~statistical point of view, \PCE\ is a~simple linear regression model with intercept. Therefore, it is possible to use \emph{ordinary least squares} (OLS) regression to solve for the coefficients $\bbeta$ that minimize the error  $ \varepsilon $.
To solve this regression problem for $\bbeta$, we first generate $ \Ns $ realizations of the input random vector $ \X $ and compute the corresponding results of the model $ \pazocal Y  $, together called the experimental design (ED) or training data set. Then, the vector of $P$ deterministic coefficients $\bbeta $ can be determined by OLS as
\begin{equation}
    \bbeta
    =
    (\bPsi^{T}\bPsi)^{-1} \ \bPsi^{T}  \pazocal Y ,
\label{Eq: PCe_OLS}
\end{equation}
where $ \bPsi $ is the data matrix
\begin{equation}
    \bPsi
    =
    \left\{
        \Psi_{ij}= \Psi_{j}(\xxi^{(i)}),  \;
        i=1, \ldots, \Ns,   \;
        j=0, \ldots, P-1
    \right\}.
\label{Eq: basis_matrix}
\end{equation}

It is clear from Eq.~\eqref{Eq.: Cardinality PCE} that $ P $ is strongly dependent on the number of input random variables $ M $ and the maximum total degree of polynomials $ p $. Considering that estimating $\bbeta$ by OLS requires at least $\mathcal{O}( P \, \ln (P))$ samples for a stable solution \cite{CohenOptimalWLS, NarayanOptimalWLS}, the problem can become computationally demanding in cases of large or strongly non-linear models. To reduce computational expense, one can use advanced model selection algorithms such as Least Angle Regression (LAR) \cite{LARS,BLATMANLARS}to find an optimal set of \PCE\ terms, and thus reduce the number of samples needed to compute the unknown coefficients. Further reduction can be obtained by incorporating additional physical constraints.

\section{Physically Constrained Polynomial Chaos Expansion (\PC{}) -- Deterministic Formulation}

In this section, we propose a novel approach to perform \PCE\ regression with known physical constraints. The approach, referred to as Physically Constrained Polynomial Chaos Expansion (\PC{}), expands the classes of regression models that can obey physical constraints to include the \PCE, which is widely used for UQ. We begin by formulating the constrained regression problem for the solution of deterministic PDEs.

Consider the general partial differential equation given by
\begin{equation}
    \begin{aligned}
        & \mathcal{L}(\bm{x},t; u(\bm{x},t)) = f(\bm{x},t), & \forall \bm{x}\in \mathcal{D}, t \in \mathcal{T}\\
        & \mathcal{B}(\bm{x},t; u(\bm{x},t)) = g(\bm{x},t,), & \forall \bm{x}\in \partial\mathcal{D}, t \in \mathcal{T} 
        \label{eqn:PDE determ}
    \end{aligned}
\end{equation}
where $\mathcal{T}\subset \mathbb{R}$, $\mathcal{D}\subset \mathbb{R}^3$ with boundary $\partial\mathcal{D}$, $\mathcal{L}$ is a differential operator with boundary operator $\mathcal{B}$, $u(\cdot)$ is the response of the system, and $f,g$ are external forces/source terms. We aim to solve the \PCE\ regression problem described above as constrained by the general Eq.~\eqref{eqn:PDE determ}. That is, we define the objective function by
\begin{equation}
    \begin{aligned}
        & \mathcal{M}(\bbeta) = \min_{\bbeta} \sum_{j=1}^{n_{\mathrm{sim}}} \left[y^j - u^{\mathrm{PCE}}(\bm{x}^j, t^j) \right]^2=  \min_{\bbeta} \lVert \pazocal{Y}  - \bPsi\bbeta \rVert^2\\
        \text{s.t. } &  \mathcal{L}(\bm{x}_{\mathrm{V}},t_{\mathrm{V}}; u(\bm{x}_{\mathrm{V}},t_{\mathrm{V}})) = f(\bm{x}_{\mathrm{V}},t_{\mathrm{V}}), \\
        & \mathcal{B}(\bm{x}_{\mathrm{BC}},t_{\mathrm{BC}}; u(\bm{x}_{\mathrm{BC}},t_{\mathrm{BC}})) = g(\bm{x}_{\mathrm{BC}},t_{\mathrm{BC}},)
    \end{aligned}
    \label{Eq. PC2 determ definition}
\end{equation}
where we define a discrete sets of $n_{\mathrm{BC}}$ samples $(\bm{x}_{\mathrm{BC}}, t_{\mathrm{BC}})$ and $n_{\mathrm{V}}$ samples $(\bm{x}_{\mathrm{V}}, t_{\mathrm{V}})$ 
 -- referred to as boundary points and virtual points, respectively -- at which the boundary conditions and PDE will be explicitly satisfied. Although $(\bm{x},t)$ is a deterministic vector, for the \PCE\ regression here it is assumed to be a random vector with uniform distribution $\pazocal{U} $ for construction of \PC{}. This is important primarily because it dictates that we use Legendre polynomials as basis functions for both space and time according to the Wiener-Askey scheme. Consequently, all deterministic input variables $(\bm{x},t)$ must be transformed to standardized space, i.e.\ $\xxi\sim \pazocal{U}\left[-1,1\right]$, via an operator $\pazocal{T}$. That is,  we define the boundary points and virtual points by $\xxi_{\mathrm{BC}}=\pazocal{T}(\bm{x}_{\mathrm{BC}}, t_{\mathrm{BC}})$ and $\xxi_{\mathrm{V}}=\pazocal{T}(\bm{x}_{\mathrm{V}}, t_{\mathrm{V}})$, respectively.

\subsection{Physical Constraints \& Lagrange Multipliers}

The constrained optimization problem defined by Eq. \ref{Eq. PC2 determ definition} can be solved efficiently using the method of Lagrange multipliers. The Lagrangian function takes the following form:
\begin{equation}
    L(\bbeta,\boldsymbol\lambda)=\frac{1}{2}\mathcal{M}(\bbeta) + \sum_{b=1}^{n_{\mathrm{BC}}}\lambda_b(a_b^{\mathrm{T}}\bbeta-c_b)+ \sum_{v=1}^{n_{\mathrm{V}}}\lambda_v(a_v^{\mathrm{T}}\bbeta-c_v)
    \label{Eq. Lagrangian}
\end{equation}
Boundary conditions are prescribed by their type defined by boundary operator $\mathcal{B}$, coordinates $\xxi_{\mathrm{BC}}$ and corresponding vector $\mathbf{c}_{\mathrm{BC}}$ consisting of $n_{\mathrm{BC}}$ rows $c_b=g(\xxi_{\mathrm{BC}}^{(b)})$.  Similarly, ODE/PDE constraints are given by the differential operator $\mathcal{L}$, virtual points $\xxi_{\mathrm{V}}$, and corresponding vector $\mathbf{c}_{\mathrm{V}}$ consisting of $n_{\mathrm{V}}$ rows $c_v=u(\xxi_{\mathrm{V}}^{(v)})$.
We then assemble these constraints into a matrix $\mathbf{A}$ where the $a_b=\{a_b^j=\mathcal{B}(\Psi_j(\xxi_{\mathrm{BC}}^{(b)})), \quad j=0,...,P-1\}$ form the first $n_{\mathrm{BC}}$ rows and vectors $a_v=\{a_v^j=\mathcal{L}(\Psi_j(\xxi_{\mathrm{V}}^{(v)})), \quad j=0,...,P-1\}$ then form the remaining $n_{\mathrm{V}}$ rows of the matrix.  Therefore, instead of solving the coefficients by $\bbeta$ by Eq. \ref{Eq: PCe_OLS}, we construct the following system of linear equations reflecting the OLS solution with physical constraints obtained from the Karush–Kuhn–Tucker (KKT) conditions:
    \begin{equation}
        \underbracket[0.4pt]{\begin{bmatrix}
        \boldsymbol\Psi^{T}\boldsymbol\Psi & \mathbf{A}^{T} \\
        \mathbf{A} & \boldsymbol{0}  
        \end{bmatrix}}_{\mathrm{KKT \; matrix }}
        \begin{bmatrix}
        \bbeta  \\
        \boldsymbol\lambda
        \end{bmatrix}
        =
        \begin{bmatrix}
        \boldsymbol\Psi^{T}  \pazocal Y  \\
        \mathbf{c}
        \end{bmatrix}
        \label{Eq. KKT system}
    \end{equation}

Construction of the $\mathbf{A}$ matrix requires derivatives of the \PCE\ model. We notice, however, that the \PCE{} and its derivatives have the same coefficients $\bbeta$ but different basis functions. The constraints can therefore be imposed by computing the derivatives of the \PCE{}, which can be done efficiently through term-wise derivatives of the basis functions as follows:
\begin{equation}
\frac{\partial^{n} f}{\partial x_i^n}=\frac{\partial^{n} \left[ \sum_{\balpha \in \pazocal A}
    \beta_{\balpha} \Psi_{\balpha}(\xxi)\right ]}{\partial \xi_i^n} \Delta_\Gamma^n=\sum_{\balpha \in \pazocal A} \beta_{\balpha} \frac{\partial^{n}\Psi_{\balpha}(\xxi)}{\partial \xi_i^n} \Delta_\Gamma^n
\end{equation}
where $\Delta_\Gamma$ reflects the different size of the time-space variable $x_i$ and standardized $\xi_i$, i.e. $\Delta_\Gamma=2/(x_{max}-x_{min})$ for Legendre polynomials defined on $\xi_i\in\left[-1,1\right]$ orthonormal to $X_i\sim\pazocal{U}\left[x_{min},x_{max}\right]$. Note that the number of terms in the \PCE{} constraints (columns of $\mathbf{A}$) is identical to the number in the original \PCE{} and it is also possible to do basic arithmetic operations on the basis to satisfy the prescribed $\mathcal{L}$ using the \PCE\ alone (i.e.\ we do not need to solve the PDEs). 
Although it is possible to generate an arbitrary number of virtual points (similar to PINNs), it leads to an over-determined system. Therefore, we use the optimal number obtained as  $n_{\mathrm{V}}=P-n_{\mathrm{BC}}$ which assures a well-determined KKT system of equations. The total computational cost of the  KKT system solution is $\pazocal{O} \left((n_{\mathrm{V}}+n_{\mathrm{BC}}+P)^3\right)$. However, additional information in the form of physical constraints incorporated in \PC{} significantly reduces $n_{\mathrm{sim}}$, which could be crucial for costly mathematical models.  

\subsection{Approximation Error Estimation}
Once the \PCE\ is constructed, it is crucial to estimate its accuracy. Further, the \PCE\ accuracy can be used to directly compare several \PCE{}s in a sparse adaptive framework to choose the best surrogate model. Ideally, the ED should be divided into validation and training sets, but this might be extremely computationally demanding in engineering applications with complex numerical models. Therefore, in the field of UQ it is often preferred to estimate the approximation error directly from the training set, without additional sampling. A~common choice of error measure is the mean squared error $ R^2 $, which is well-known from machine learning and statistics. However, driving $ R^2 $ down may lead to over-fitting.  One of the most widely-used methods in UQ is the leave-one-out cross-validation (LOO-CV) error $ Q^2 $, which can be obtained analytically from a~single \PCE\ \cite{BLATMAN2010}. However, since the \PC{} is designed for small ED, $ Q^2 $ may not be recommended and $ R^2 $ is preferred in most cases, possibly normalized by the variance of the QoI as commonly used in surrogate modeling.  The total mean squared approximation error consists of three terms -- error in the approximated function $ R^2_u $, error associated with failure to obey the PDE constraints $ R^2_{\mathcal{L}}$, and error in the boundary conditions $ R^2_{\mathcal{B}}$, formally written as:
\begin{equation}
    R^2
    =  R^2_u + R^2_{\mathcal{L}} + R^2_{\mathcal{B}}
    \label{Eq: RR}
\end{equation}
The tendency of $ R^2 $ to over-fitting is mitigated by the second and third components of the error measure -- the mean squared error in the given PDE and boundary conditions.

\section{\PC{} for Uncertainty Quantification}
\label{UQ}

In this section, we extend the deterministic formulation of \PC{} to perform UQ -- which is the essential motivation for using \PCE{} as opposed to other physically constrained ML methods such as PINNs. Consider the general stochastic partial differential equation given by
\begin{equation}
    \begin{aligned}
        & \mathcal{L}(\bm{x},t,\X(\omega); u(\bm{x},t,\X(\omega))) = f(\bm{x},t,\X(\omega)), & \forall \bm{x}\in \mathcal{D}, t \in \mathcal{T}, \omega \in \Omega\\
        & \mathcal{B}(\bm{x},t,\X(\omega); u(\bm{x},t,\X(\omega))) = g(\bm{x},t,\X(\omega)), & \forall \bm{x}\in \partial\mathcal{D}, t \in \mathcal{T}, \omega \in \Omega
        \label{eqn:PDE UQ}
    \end{aligned}
\end{equation}
where the  meaning of the symbols is identical to Eq. \eqref{eqn:PDE determ} and $\X(\omega)\in \mathbb{R}^d$ is a $d$-dimensional random vector having sample space $\Omega$. Now, we aim to solve the \PCE\ regression problem described above as constrained by the general Eq.~\eqref{eqn:PDE UQ}. That is, we define the objective function as
\begin{equation}
    \begin{aligned}
        & \mathcal{M}(\bbeta) = \min_{\bbeta} \sum_{j=1}^{n_{\mathrm{sim}}} \left[y^j - u^{\mathrm{PCE}}(\bm{x}^j, t^j, \X^j) \right]^2=  \min_{\bbeta} \lVert \pazocal{Y}  - \bPsi\bbeta \rVert^2\\
        \text{s.t. } &  \mathcal{L}(\bm{x}_{\mathrm{V}},t_{\mathrm{V}},\X(\omega); u(\bm{x}_{\mathrm{V}},t_{\mathrm{V}},\X(\omega))) = f(\bm{x}_{\mathrm{V}},t_{\mathrm{V}},\X(\omega)), \\
        & \mathcal{B}(\bm{x}_{\mathrm{BC}},t_{\mathrm{BC}},\X(\omega); u(\bm{x}_{\mathrm{BC}},t_{\mathrm{BC}},\X(\omega))) = g(\bm{x}_{\mathrm{BC}},t_{\mathrm{BC}},\X(\omega))
    \end{aligned}
    \label{Eq. PC2 UQ definition}
\end{equation}
where we define a discrete sets of $n_{\mathrm{BC}}$ boundary points $(\bm{x}_{\mathrm{BC}}, t_{\mathrm{BC}},\X)$ and $n_{\mathrm{V}}$ virtual points $(\bm{x}_{\mathrm{V}}, t_{\mathrm{V}},\X)$. As in the deterministic formulation, all input variables are transformed to standardized space according to the Wiener-Askey scheme, i.e. $\xxi_{\mathrm{BC}}=\pazocal{T}(\bm{x}_{\mathrm{BC}}, t_{\mathrm{BC}},\X)$ and $\xxi_{\mathrm{v}}=\pazocal{T}(\bm{x}_{\mathrm{V}}, t_{\mathrm{V}},\X)$. Note that the boundary and virtual points retain the random vector $\X$, and do not (necessarily) contain specific boundary or virtual points for $\X$. This is because physical constraints are typically expressed in terms of the physical variables $(\bm{x},t)$ and are not (in general) expressed in terms of the random variables contained in $\X$. That is, $\X$ does not necessarily affect the constraints associated with either $\mathcal{L}$ and $\mathcal{B}$ (although it can). 
As a result, the random vector $\X$ usually serves only to increase the dimension of the input random vector and thus the solution of the Eq. \eqref{Eq. PC2 UQ definition} follows the same form as the deterministic formulation given in Eq. \eqref{Eq. KKT system}.

If, on the other hand, the physical constraints associated with the PDE or BCs are expressed in terms of the random variables in $\X$, virtual and/or boundary points can be used to enforce these constraints in the same manner as performed above. If these are expressed as equality constraints, the KKT equations to solve the constrained optimization do not change. On the other hand, if these constraints are expressed as inequalities (e.g. non-negative coefficients) then more advanced optimizers are required. This will be the topic of a future work.

The resulting constrained \PCE{} is expressed as a function of the complete set of physical and random variables $\xxi =\pazocal{T}(\bm{x}, t,\X)$. The form of the \PCE\ as a~linear summation over orthonormal polynomials allows for powerful and efficient post-processing. In particular, once a~\PCE\ approximation is created, it is possible to directly estimate statistical moments of the output from the expansion. The first statistical moment (the mean value) is simply the first deterministic coefficient of the expansion $\mu_{Y}=\big<Y^{1} \big> = \beta_{\bzero}$. The second raw statistical moment, $ \big<Y^{2} \big> $, can be estimated by
\begin{align}
    \label{Eq:SecondRawMoment}
    \left\langle {{Y^2}} \right\rangle
    &=
    \int
    {\left[ {\sum\limits_{\balpha \in \A}
        {{\beta _{\balpha}}{\Psi _{\balpha}}\left( \xxi  \right)} } \right]} ^2
        p_{\xxi} \left( \xxi \right)   \;
        \mathrm{d}  \xxi =
    \sum\limits_{\balpha_1 \in \A}
    \sum\limits_{\balpha_2 \in \A}
    \beta_{{\balpha}_1}
    \beta_{{\balpha}_2}
    \int
    \Psi_{{{\balpha}_1}}\left( \xxi \right)
    \Psi_{{{\balpha}_2}}\left( \xxi \right)
    p_\xxi \left( \xxi \right)  \;
    \mathrm{d} \xxi
    \\   \nonumber
    & =
    \sum\limits_{{\balpha} \in {\A}} {\beta_{\balpha}^2} {\int {{\Psi _{\balpha}}\left( \xxi \right)} ^2}{p_\xxi}
    \left( \xxi\right)  \;
    \mathrm{d}  \xxi =
    \sum\limits_{{\balpha} \in {\A}} {\beta_{\balpha}^2} \left\langle {{\Psi _{\balpha}},{\Psi _{\balpha}}} \right\rangle = \sum\limits_{{\balpha} \in {\A}} {\beta_{\balpha}^2}.
\end{align}
Considering the orthonormality of the polynomials, it is possible to obtain the variance  $ 	\sigma_{Y}^{2}=\big<Y^{2} \big>- \mu_{Y}^{2} $
 as the sum of all squared deterministic coefficients except the intercept (which represents the mean value). Note that the estimation of higher statistical central moments, specifically skewness and kurtosis, are more complicated since they require triple and quad products. These can be obtained analytically only for certain polynomial families, e.g. formulas for Hermite and Legendre polynomials (and their combination) can be found in \cite{NOVAK2022106808}. Moreover, it can be shown that \PCE{} is in the form of the Hoeffding-Sobol decomposition and thus Sobol indices can be obtained analytically by applying Eq. \eqref{Eq:SecondRawMoment} to selected subsets of the \PCE{} terms \cite{SUDRET}.

As can be seen, the main advantage of \PCE{} over other surrogate models lies in the convenience of performing uncertainty quantification, i.e. the above analytical moment and sensitivity estimates. The \PC{} framework inherits this convenience when appropriate conditions are established. As noted, the \PCE{} is expressed in terms of both the random variables $\X$ and the deterministic space-time coordinates $(\bm{x},t)$.
Therefore, 
to properly estimate statistical moments or sensitivity indices, we must condition the \PCE{} on $(\bm{x},t)$, yielding  local space-time statistics. 
The numerical estimation of local mean value $\mathbb{E}\left[f\rvert \bm{x},t \right]$ and variance $\sigma^2_{\left[f\rvert \bm{x},t \right]}$ are based on a simple rationale: deterministic values can be specified directly as constants in the \PCE{} terms. Following the concept of reduced \PCE{} \cite{NOVAK2022106808}, the local variance $ \sigma^2_{\left[f\rvert x,t \right]}$ is affected only by \PCE{} terms containing the random variables $\X$, while all other terms are constants whose sum is equal to a local expected value $\mathbb{E}\left[f\rvert x,t \right]$.

Formally stated, the random variables form a subset of all $M$ input variables $\{\bm{x},t, \X\}$, i.e., $\X\subset \{\bm{x},t, \X\}$. We thus define the subset $ \mathbf{u} \subseteq \mathrm{I}= \{1,...,M\} $ and its complement $  {\bf{u}^c} $ as

\begin{equation}
\mathbf{u} = \left\{ \mathrm{i} \in \mathrm{I}: \xi_i \in \pazocal{T}(\X) \right\}, \qquad \mathbf{u}^c = \left\{ \mathrm{i} \in \mathrm{I}: \xi_i \in \pazocal{T}(\{\bm{x},t\}) \right\}.
\end{equation}
For the sake of clarity, the set of basis multivariate polynomials dependent on $ \X$ and its complement are

\begin{equation}
{\pazocal{A}_{\X}} = \left\{ {{\bf{\alpha }} \in {\pazocal{A}}:{\alpha _k} \ne 0 \leftrightarrow k \in {\bf{u}}\;} \right\}, \qquad {\pazocal{A}_{\sim\X}} = \left\{ {{\bf{\alpha }} \in {\pazocal{A}}:{\alpha _k} \ne 0 \leftrightarrow k \in {\bf{u}^c}\;} \right\}.
\end{equation}

The local variance is further obtained by a simple post-processing of the terms containing random variables. First, it is necessary to find a unique basis $\pazocal{A}_{\X}$ -- defined by the unique set of polynomials from $\pazocal{A}_{\X}$. Although the original $\pazocal{A}$ contains only unique terms, after the reduction there could be identical terms, e.g. two terms differing only in polynomial degrees associated to variables in $ {\bf{u}^c}  $. Further, coefficients of terms with duplicate multi-indices are summed and assigned to the unique term in the reduced \PCE{}. The constants arising in mixed \PCE{} terms containing deterministic and random variables must be added to coefficients associated to the unique basis. Finally, local mean and variance are obtained by commonly known formulas directly from coefficients of the reduced \PCE{}, i.e. expected value from the coefficients associated with terms containing only deterministic variables in complement set $ \pazocal{A}_{\sim\X}  $ and variance from coefficients associated to terms in $\pazocal{A}_{\X}$:

\begin{align}
    \mathbb{E}\left[u\rvert \bm{x},t \right] &= \beta_0 +  \sum\limits_{\balpha  \in {\pazocal{A}_{\sim \X}}} \beta_{\balpha}\\
    \sigma^2_{\left[u\rvert \bm{x},t \right]} &=  \sum\limits_{\balpha  \in {\pazocal{A}_{\X}}} \beta^2_{\balpha}
\end{align}

\section{Numerical Algorithms}
In this section we present computationally efficient means to solve the proposed approach for the \PC{} coefficients. We then extend the approach to use a reduced basis set defined through LAR. Finally, an iterative scheme is proposed to fit the \PC{} constrained by nonlinear PDEs. Implementation of the proposed approach is simple and straightforward: instead of a common solution by OLS, one can use the KKT system in existing non-intrusive regression-based algorithms. Therefore, it is possible to use various adaptive sparse non-intrusive algorithms for the construction of PCE. The main three algorithms are presented in the following paragraphs.

The basic algorithm for the construction of the KKT normal equations builds sub-matrix $\mathbf{A}$ containing the given physical constraints in the form of boundary/initial conditions evaluated at $\xxi_{\mathrm{BC}}$ and PDE evaluated at $\xxi_{\mathrm{V}}$. The sub-matrix $\mathbf{A}$ prescribes additional constraints and initial conditions to $\bbeta$ leading to efficient and accurate estimations for very low-size (or even missing) ED. Finally,  the unknown vector $\bbeta$ is obtained by OLS as a solution of the KKT matrix composed of $\mathbf{A}$ and the information sub-matrix $\boldsymbol\Psi^{T}\boldsymbol\Psi$ prepared according to Eq. \ref{Eq. KKT system}. The whole algorithm is detailed in Algorithm \ref{Alg: KKT}.

\begin{algorithm}[!ht]
	\caption{KKT solver with virtual samples}
\begin{algorithmic}[1]
\Statex	\textbf{Input:}  ED ($\{\bm{x},t, \X\}$ and $\pazocal Y$), $\boldsymbol\Psi$, boundary conditions ($\boldsymbol{\xi}_{\text{\tiny BC}}$, $\mathbf{c}_{\mathrm{BC}}$), $\mathcal{L}$, $\mathcal{B}$, $f$
\For{$b \gets 1$ to $n_{\mathrm{BC}}$}
\For{$j \gets 1$ to $P$}

\State evaluate $a_b^j=\mathcal{B}(\Psi_j(\xxi_{\mathrm{BC}}^{(b)}))$
\State	add $a_b^j$ to $\mathbf{A}$
\EndFor
\EndFor

\State sample $n_{\mathrm{V}}=P-n_{\text{\tiny BC}}$ virtual points $\xxi_{\mathrm{V}}$ (MC, LHS etc.)
\For{$v \gets 1$ to $n_{\mathrm{V}}$}

\For{$j \gets 1$ to $P$}
\State evaluate $a_v^j=\mathcal{L}(\Psi_j(\xxi_{\mathrm{V}}^{(v)}))$
\State	add $a_v^j$ to $\mathbf{A}$
\EndFor
\State evaluate $c_v=f(\xxi_{\mathrm{V}}^{(v)})$ 
\State	add $c_v$ to $\mathbf{c}_{\mathrm{V}}$ 
\EndFor
\State assemble vector $\mathbf{c}=\left[\mathbf{c}_{\mathrm{BC}},\mathbf{c}_{\mathrm{V}} \right]$
\State construct KKT normal equations according to Eq. \ref{Eq. KKT system}
\State solve the system by OLS
             
\Statex	\textbf{Output:} PCE coefficients $\bbeta $
                        
\end{algorithmic}
\label{Alg: KKT}
\end{algorithm}

 Since the KKT-based estimation of the \PCE{} coefficients $\bbeta$ is computationally efficient, Algorithm \ref{Alg: KKT} can be easily combined with adaptive sparse solvers such as LAR. For the sake of clarity, the whole algorithm combining LAR and KKT solver with virtual samples is provided in Algorithm \ref{Alg: LAR-KKT}.

\begin{algorithm}[!ht]
	\caption{LAR-KKT algorithm}
\begin{algorithmic}[1]
\Statex	\textbf{Input:}   ED ($\{\bm{x},t, \X\}$ and $\pazocal Y$), $\boldsymbol\Psi$, boundary conditions ($\boldsymbol{\xi}_{\text{\tiny BC}}$, $\mathbf{c}_{\mathrm{BC}}$), $\mathcal{L}$, $\mathcal{B}$, $f$, $p_{\mathrm{max}}$
\For{$p \gets 1$ to $p_{\mathrm{max}}$} 
\State generate set of basis functions $\A$ 
\State identify the sequence of the most important basis functions by LAR 
\For{$i \gets 1$ to $P$}
\State construct  $\boldsymbol\Psi$
\State estimate $\bbeta $ by KKT (Algorithm \ref{Alg: KKT})
\State get approximation error $R^2_{\mathrm{LAR}}$
\State Optional: Check over-fitting 
\EndFor
\EndFor
\Statex	\textbf{Output:} $\{\boldsymbol\Psi,\bbeta  \}$ associated to the lowest $R^2_{\mathrm{LAR}}$ 
                        
\end{algorithmic}
\label{Alg: LAR-KKT}
\end{algorithm}

 \begin{figure}[h!]
\centering
	\includegraphics[width=0.65\textwidth]{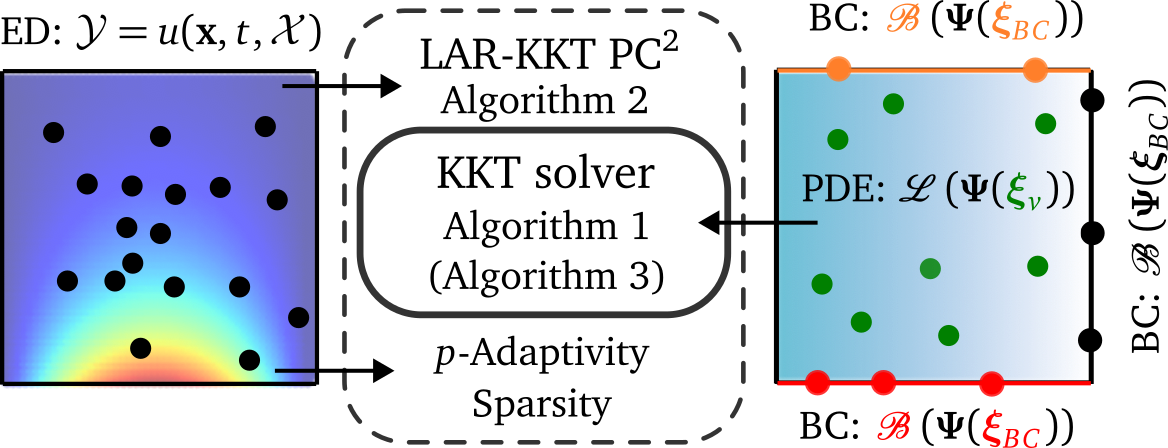}
 \caption{Graphical representation of LAR-KKT algorithm combining $p$-adaptivity, LAR algorithm and KKT solver with virtual samples.}
\label{Fig. KKT-LAR scheme}	
\end{figure}

The LAR-KKT algorithm combines information from the ED and physical constraints to identify the optimal form of the \PCE{} approximation as schematically depicted in Fig. \ref{Fig. KKT-LAR scheme}. Information obtained from the ED is specifically utilized to identify the most suitable basis set by standard LAR. Physical constraints are further used to estimate $\bbeta$ satisfying the given constraints and a suitable error measure $R^2$ is then used for the selection of the most accurate \PC{}. However, the KKT solver can be also used without Algorithm 2, which dramatically decreases the computational cost of \PC{} for low to moderate dimensions of the input space. Note that the KKT solution can be generally combined with any existing adaptive sparse regression algorithm by simply replacing standard OLS with the solution of the KKT system.

The solution is more complicated for non-linear PDEs since it is necessary to know the complete form of the \PCE{} and thus $\bbeta$. Naturally, the \PCE{} and its derivatives evaluated at the virtual samples are not known at the start of the algorithm. Therefore, it is necessary to construct the \PCE{} (estimate coefficients $\bbeta$) without virtual samples in the first step and iteratively improve the estimate by evaluation of the PDE for the virtual samples. The KKT algorithm for simple non-linear PDEs is presented in Algorithm \ref{Alg: LAR-KKT_iterative}.

\begin{algorithm}[!ht]
	\caption{Iterative KKT solver for non-linear PDEs}
\begin{algorithmic}[1]
\Statex	\textbf{Input:}   ED ($\{\bm{x},t, \X\}$ and $\pazocal Y$), $\boldsymbol\Psi$, boundary conditions ($\boldsymbol{\xi}_{\text{\tiny BC}}$, $\mathbf{c}_{\mathrm{BC}}$), $\mathcal{L}$, $\mathcal{B}$ 

\State get $\mathbf{c}_{\mathrm{BC}}$ and ${\bbeta }_0$ from Algorithm \ref{Alg: KKT} without  $\mathcal{L}$

\State sample $n_{\mathrm{V}}=P-n_{\text{\tiny BC}}$ virtual points $\xxi_{\mathrm{V}}$ (MC, LHS etc.)
\For{$v \gets 1$ to $n_{\mathrm{V}}$}
\For{$j \gets 1$ to $P$}

\State evaluate $a_v^j=\mathcal{L}_{{\bbeta }_0}(\Psi_j(\xxi_{\mathrm{V}}^{(v)}))$
\State	add $a_v^j$ to $\mathbf{A}$
\EndFor
\State evaluate $c_v=f(\xxi_{\mathrm{V}}^{(v)})$ 
\State	add $c_v$ to $\mathbf{c}_{\mathrm{V}}$ 
\EndFor
\State assemble vector $\mathbf{c}=\left[\mathbf{c}_{\mathrm{BC}},\mathbf{c}_{\mathrm{V}} \right]$
\For{$i \gets 1$ to $n_{\mathrm{iter}}$}
\State construct KKT normal equations
\State get $\bbeta_i $ by least squares
\For{$v \gets 1$ to $n_{\mathrm{V}}$}

\For{$j \gets 1$ to $P$}
\State evaluate $a_v^j=\mathcal{L}_{{\bbeta_i }}(\Psi_j(\xxi_{\mathrm{V}}^{(v)}))$
\State	add $a_v^j$ to $\mathbf{A}$
\EndFor

\State evaluate $c_v=u(\xxi_{\mathrm{V}}^{(v)})$ 
\State	add ${c}_v$ to $\mathbf{c}_{\mathrm{V}}$ 
\EndFor
\EndFor            
\Statex	\textbf{Output:} PCE coefficients $\bbeta = \bbeta_{n_{\mathrm{iter}}} $
                        
\end{algorithmic}
\label{Alg: LAR-KKT_iterative}
\end{algorithm}

\section{Numerical Experiments}
To present the capabilities of the proposed \PC{} approach, several examples are presented for various types of PDEs. First, \PC{} is applied to three 1D examples of different nature, including an inhomogeneous ODE, an ODE with BCs of arbitrary order, and a non-linear ODE. \PC{} is then used to approximate the solution to the 2D wave equation PDE and the heat equation PDE with an uncertain input variable. \PC{} is constructed using Algorithm 1 (KKT) and the sparse adaptive Algorithm 2 using KKT (LAR-KKT) and the results are compared to the standard unconstrained LAR method implemented in UQPy \cite{olivier2020uqpy,tsapetis2023uqpy}.  All compared methods are extended by $p$ adaptivity ($p\in\left[5,25\right]$) governed by $R^2$ measure to obtain the highest possible accuracy. The ED is generated by Latin Hypercube Sampling (LHS) and virtual samples are generated by Crude Monte Carlo (MC) unless stated otherwise. The number of virtual samples is $n_{\mathrm{V}}=P-n_{\mathrm{BC}}$. Note that Dirichlet BCs are incorporated in the ED for LAR method for a fair comparison, though additional information from the PDE and higher-order BCs can only be incorporated by \PC{}. All numerical results were replicated for 100 independent trials and statistical results are presented in convergence plots showing the mean $\pm\sigma$ interval. The convergence is measured by two global quantities characterizing the whole input domain: mean squared error (Eq. \ref{Eq: Error measure mean}) and maximum squared error (Eq. \ref{Eq: Error measure max}). Moreover, detailed insight is given also by a local measure represented by the absolute difference between the original model $u$ and its approximation $\tilde{u}$ (Eq. \ref{Eq: Error measure abs}). For numerical evaluation of $R^2_u$ and $R^2_{\mathcal{L}}$ we use a validation set containing $1000^M$ samples generated by MC together with $100$ samples for $R^2_{\mathcal{B}}$.

\begin{align}
  \epsilon_{\mathrm{mean}} \coloneqq & \mathbb{E}\Big[\big(u(\bm{x},t, \X)-\tilde{u}(\bm{x},t, \X)\big)^2\Big]            
  \label{Eq: Error measure mean}\\
    \epsilon_{\mathrm{max}} \coloneqq & \max\Big[\big(u(\bm{x},t, \X)-\tilde{u}(\bm{x},t, \X)\big)^2\Big]            
  \label{Eq: Error measure max}\\
    \lvert\epsilon \rvert \coloneqq & \lvert u(\bm{x},t, \X)-\tilde{u}(\bm{x},t, \X)\rvert     
  \label{Eq: Error measure abs}
\end{align}

\subsection{1D Poisson Equation: Inhomogeneous ODE}
The very first example shows the application of the \PC{} to simple 1D inhomogeneous ODE. Homogeneous and inhomogeneous ODEs/PDEs solved by \PC{} differ only in vectors $\mathbf{c}_{\mathrm{V}}$ -- containing specific values for inhomogenous PDEs and zeros for homogenous PDEs. Naturally, $\mathbf{c}_{\mathrm{V}}$ could contain constants or results of functions $f(\bm{x}_{\mathrm{V}},t_{\mathrm{V}},\X(\omega))$.  The Poisson equation and assumed BC are as follows:
\begin{align}
\Delta u(x) &= 2, \qquad x \in [-1, 1],\\
\left.\dfrac{\partial u(x)}{ \partial x}\right|_{x=1} &=4, \qquad u(-1)=0. \nonumber
\end{align}
where $\Delta$ is the Laplace operator.

\begin{figure}[ht]
\centering
	\includegraphics[width=1\textwidth]{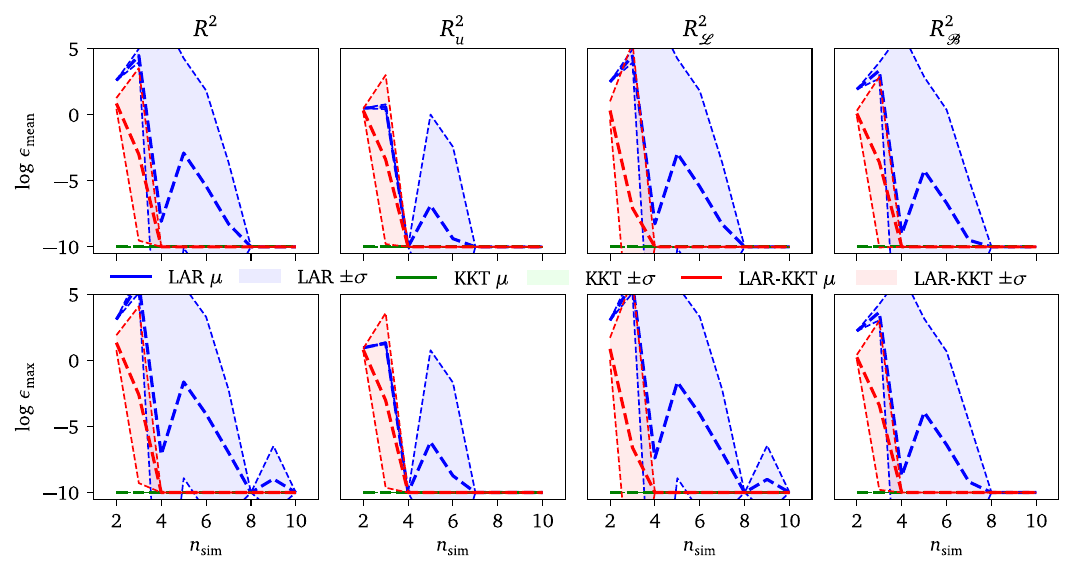}
 \caption{Obtained numerical results for 1D Poisson equation for increasing number of samples in the domain. Top row shows mean squared errors and bottom row shows maximum squared errors obtained by simple KKT (Algorithm 1), LAR-KKT (Algorithm 2) and standard LAR method. Each column corresponds to a component of total error measure $R^2$.}
	\label{Fig. Poisson}
\end{figure}

The obtained numerical results are summarized in Fig. \ref{Fig. Poisson}. Note that although the figure contains also $\pm\sigma$ intervals, these are not visible for the KKT method due to extremely low variance in the obtained accuracy. The accuracy is measured by three quantities $ R^2_u, R^2_{\mathcal{L}}, R^2_{\mathcal{B}}$ and their sum representing total error ($R^2$). Note that while $R^2_u$ is typically a sufficient measure in standard surrogate modeling, in physically informed surrogate models it is also important to measure $R^2_{\mathcal{L}}, \ R^2_{\mathcal{B}}$ to ensure the physical constraints are being satisfied. Standard LAR shows convergence to the exact solution measured by $R^2_u$ from 8 samples in the domain. The KKT method is exact even for two realizations in the domain. The KKT-LAR method combining the benefits of sparse solution and KKT approach converges in 4 samples while having large variance for 3 samples. This behavior of the method can be seen also in the next linear ODE example, where the effect of sparsity is not so significant and it is thus generally more efficient to use the standard KKT method in simple linear ODEs.  Although all compared methods converge to the exact solution for maximum number of $n_{\mathrm{sim}}=10$, standard LAR method leads to significant errors for lower $n_{\mathrm{sim}}$. Note that although LAR has almost perfect accuracy measured by $R^2_u$ for $n_{\mathrm{sim}}=4$, it still has significant errors in the remaining criteria while KKT and LAR-KKT converge consistently in total $R^2$. For the sake of completeness, the maximum errors (bottom row) complement the mean errors (top row).

\begin{figure}[h!t]
\centering
	\includegraphics[width=1\textwidth]{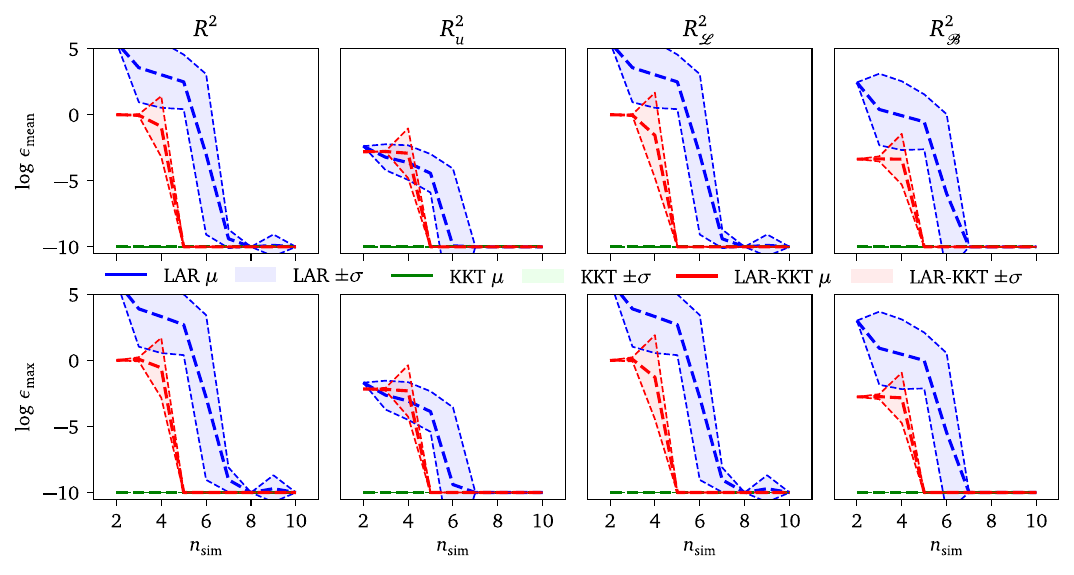}
	 \caption{Obtained numerical results for 1D Euler equation for increasing number of samples in the domain. Top row shows mean squared errors and bottom row shows maximum squared errors obtained by simple KKT (Algorithm 1), LAR-KKT (Algorithm 2) and standard LAR method. Each column corresponds to a component of total error measure $R^2$.}
  \label{Fig. Euler}
\end{figure}

\subsection{1D Euler Equation: Arbitrary BC}
\PC{} offers efficient $n$th derivatives of the basis functions and thus  is possible to solve PDES containing generally derivatives of arbitrary order in $\mathcal{L}$ and $\mathcal{B}$, if the $n$th derivative of the \PCE{} basis exists. This example presents the 1D Euler beam equation with given BCs as follows:
   \begin{align}
\frac{\partial^{4} u(x)}{\partial x^4} + 1 &= 0, \qquad x \in [0, 1],\\
u(0)=0, \qquad u'(0)&=0 \qquad u''(1)=0, \qquad  u'''(1)=0\nonumber
\end{align}
This ODE with the prescribed BCs represents the deflection of a cantilever beam with a uniformly distributed load having intensity $q=1$. The obtained results are shown in Fig. \ref{Fig. Euler} in the identical form as in the previous example. The convergence trends of the compared methods are similar to the previous example, though the differences between standard LAR and \PC{} are more significant. Consistency of \PC{} convergence with respect to all criteria can be clearly seen in convergence plots of the LAR-KKT method, while the standard LAR approach leads to similar overfitting for $n_{\mathrm{sim}}=6$ similar to the previous example.

\subsection{1D Logistic Equation: Non-linear ODE}
The last ODE example presenting capabilities of the proposed \PC{} is a non-linear ODE. Non-linear ODEs/PDEs solved by \PC{} require iterative construction of KKT system using Algorithm 3 with LHS sampling for virtual points, though it is not applicable to strongly non-linear PDEs. For more complicated PDEs, it will be necessary to use more advanced and more computationally expensive optimizers. The example is a 1D Logistic equation commonly representing population growth in the following form:
\begin{align}
    \dfrac{\partial u(x)}{\partial x}&=u(x)\left(1-u(x)\right),  \qquad x \in [-5,5],\label{Eq. ODE} \\
    \quad u(0)&=0.5\nonumber
\end{align}

\begin{figure}[b!]
\centering
	\includegraphics[width=1\textwidth]{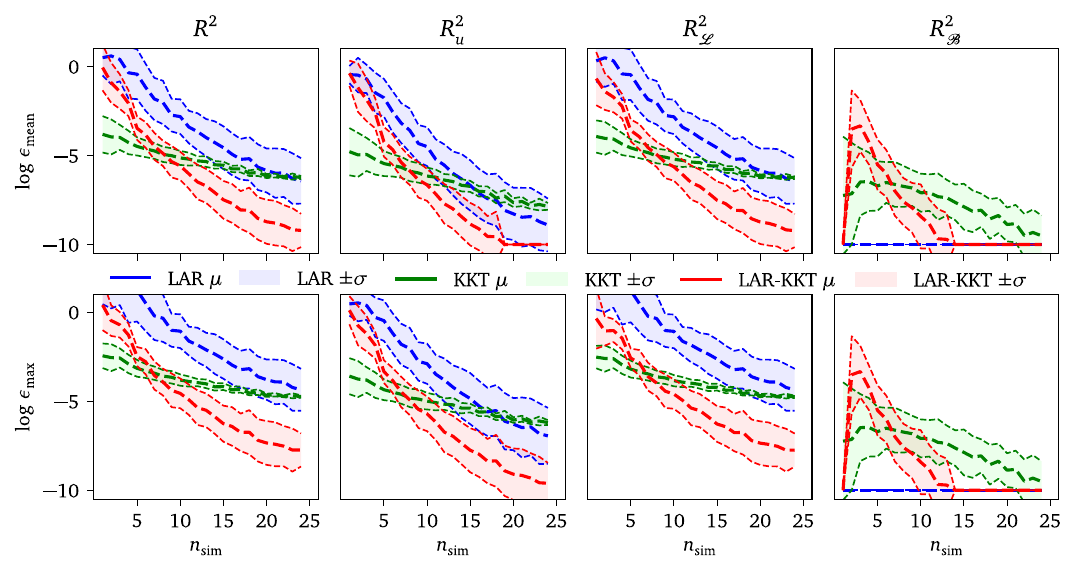}
 	 \caption{Obtained numerical results for 1D Logistic equation for increasing number of samples in the domain. Top row shows mean squared errors and bottom row shows maximum squared errors  obtained by simple KKT (Algorithm 1), LAR-KKT (Algorithm 2) and standard LAR method. Each column corresponds to a component of total error measure $R^2$.}
	\label{Fig. Logistic}
\end{figure}

The obtained numerical results are summarized in Fig. \ref{Fig. Logistic}. The convergence trend is similar for all methods, but \PC{} achieves significantly higher accuracy. Note that the simple KKT solver is the most accurate for a very low number of samples, though for increasing number of samples in the domain the benefit of the sparse solution obtained by the LAR-KKT solver leads to superior accuracy. Interestingly, the convergence trend of standard LAR is very similar to LAR-KKT though there is a significant difference in the absolute accuracy. 
Although the total accuracy of LAR is low, the accuracy in the given BC is perfect, possibly indicating that LAR is overfitting. In contrast to LAR, both \PC{} methods consistently converge in all presented criteria. Relatively slow convergence of $R_{\mathcal{L}}$ and $R_{\mathcal{B}}$ obtained by the KKT solution clearly show the necessity of sparse solvers in this example, since some basis functions associated with high $p$ are the most suitable basis of an accurate \PCE{} approximation. However, the cardinality of the full set of basis functions used in the KKT method is high, and thus it needs a high $n_{\mathrm{sim}}$ for accurate coefficient estimates.

\subsection{Wave equation: Time-dependent PDE}
The previous 1D examples clearly show possibilities of \PC{} for various types of ODEs. Naturally, the \PC{} can be used for higher number of input variables as will be presented in the time-dependent wave equation:
\begin{align}
\label{Eq. Wave Equation}\frac{\partial^2u(x,t)}{\partial t^2} &= 4\frac{\partial^2u(x,t)}{\partial x^2},  \qquad x \in [0,1], t \in [0,2] \\
\frac{\partial u(x,0)}{\partial t}&=0, \qquad u(0,t)=u(1,t) = 0, \qquad u(x,0) =\sin{(\pi x)}. \nonumber
\end{align}

\begin{figure}[b!]
\centering
	\includegraphics[width=1\textwidth]{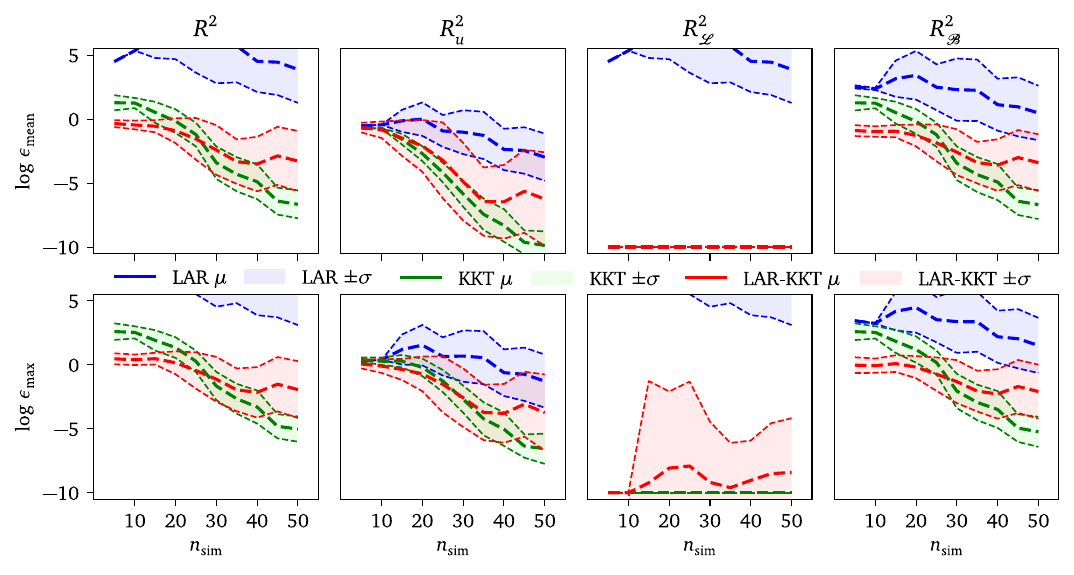}
	\caption{Obtained numerical results for wave equation for increasing number of samples in the domain. Top row shows mean squared errors and bottom row shows maximum squared errors  obtained by simple KKT (Algorithm 1), LAR-KKT (Algorithm 2) and standard LAR method. Each column corresponds to a component of total error measure $R^2$.}
  \label{Fig. Wave Equation}
\end{figure}

Although the extension of \PC{} for $n$-dimensional problems is straightforward since it is based on standard \PCE{}, it brings one significant difference in comparison to 1D examples: boundary and initial conditions are now always prescribed point-wise and thus the accuracy of \PC{} is also dependent on number and position of samples on the boundaries. In this paper, we use the DeepXDE python package \cite{lu2021deepxde} for deterministic sampling on the boundaries and thus their position does not influence the variance of the convergence plots. Convergence plots for $n_{\mathrm{BC}}=10$ samples on the boundaries are shown in Fig. \ref{Fig. Wave Equation}. \PC{} converges rapidly to very accurate approximation, although the $R_u^2$ for very low $n_{\mathrm{sim}}$ is identical to LAR. Moreover, one can see that the PDE error $R_{\mathcal{L}}^2$  is negligible for \PC{}, but LAR does not respect the PDE, and thus it does not converge to the original model. Note that the difference between \PC{} and LAR is generally significantly higher for increasing $n_{\mathrm{BC}}$  due to additional samples with Neumann BC reflected only by \PC{}.

\begin{figure}[h!]
\centering
	\includegraphics[width=0.8\textwidth]{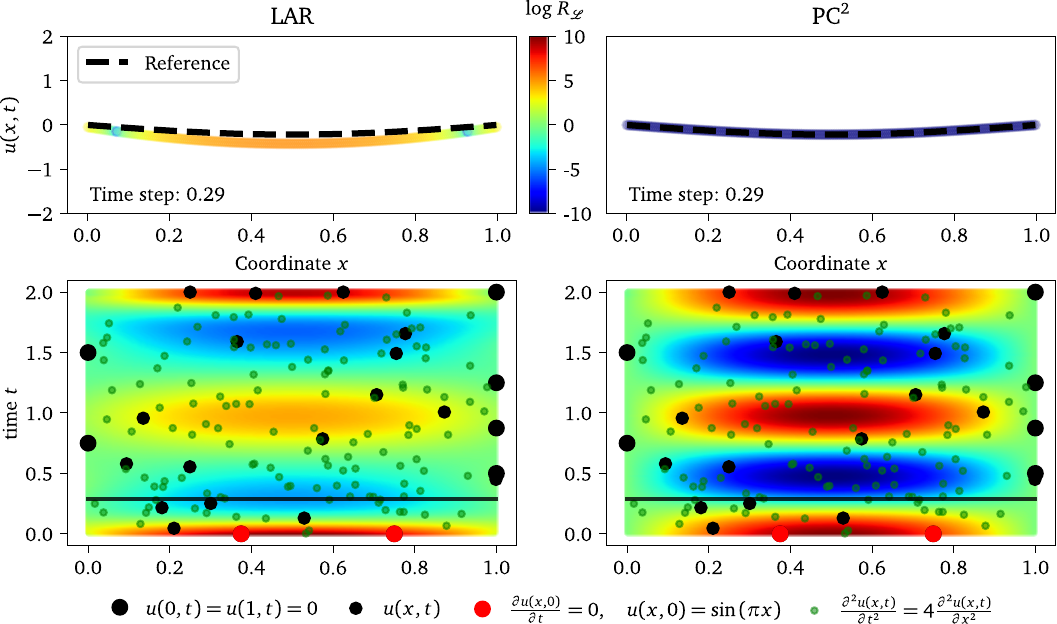}
	\caption{Selected realization of LAR (left) and \PC{} (right) for the 2D wave equation. The left pane shows significant error $R_u^2$ and $R_\mathcal{L}^2$ for standard LAR. The right pane shows a very accurate solution from \PC{} using identical ED, $\bm{x}_{\mathrm{BC}}$ and  $\bm{x}_{\mathrm{V}}$.}
  \label{Fig. Wave Equation realization 1}
\end{figure}

\begin{figure}[h!]
\centering
	\includegraphics[width=0.8\textwidth]{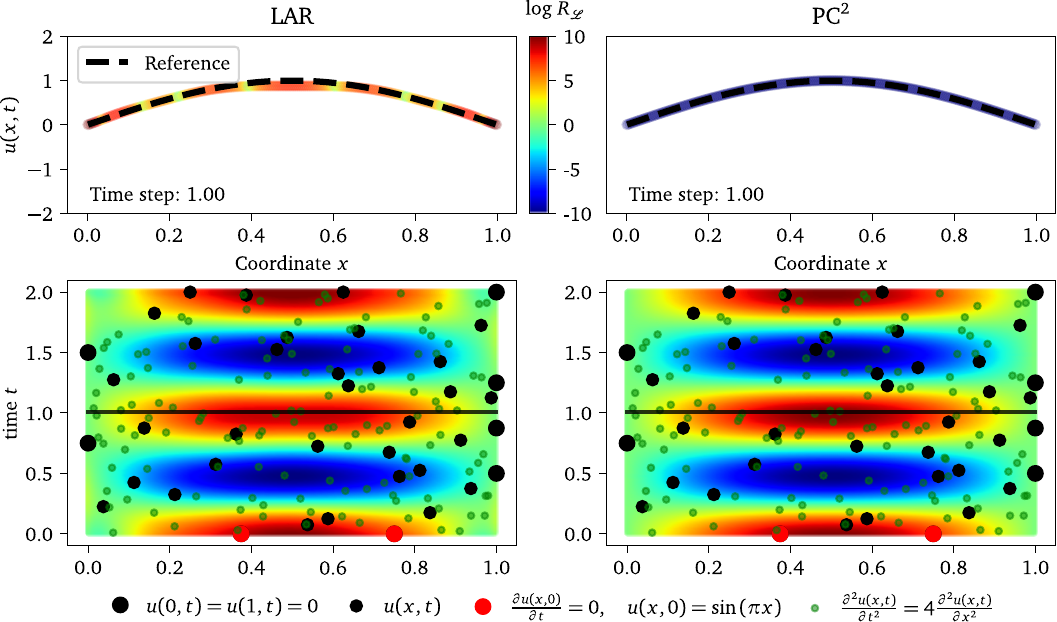}
	\caption{Selected realization of LAR (left) and \PC{} (right) for the 2D wave equation. The left and right panes show similar errors $R_u^2$ (bottom) but \PC{} shows much lower error in PDE $R_\mathcal{L}^2$ (top) using identical ED, $\bm{x}_{\mathrm{BC}}$ and  $\bm{x}_{\mathrm{V}}$.}
  \label{Fig. Wave Equation realization 2}
\end{figure}

A detailed comparison of the \PC{} and standard LAR method for a selected realization of the algorithm can be seen in Fig.\ref{Fig. Wave Equation realization 1}. The top row shows the solution of the wave equation for a selected $t$ where the dashed line represents the analytical reference solution and the colored line shows the solution by LAR (left) and \PC{} (right). The approximations are discretized to 200 points and their colors corresponds to the logarithm of the local squared errors of the PDE $\log R^2_{\mathcal{L}}$. The bottom row shows approximations of the wave equation over the entire input space together with given the ED, $\bm{x}_{\mathrm{BC}}$ and $\bm{x}_{\mathrm{V}}$. It is clear that standard LAR does not lead to sufficiently accurate approximation of $u(x,t)$, which is also reflected by high $R^2_{\mathcal{L}}$. Furthermore, even when LAR produces similar errors $R_u^2$, its PDE errors $R_{\mathcal{L}}^2$ are much larger as illustrated in Fig. \ref{Fig. Wave Equation realization 2}. In this case, LAR leads to $R<10^{-3}$ though it is clear from  $\log R^2_{\mathcal{L}}$ that the prescribed PDE is not respected. Note that \PC{} leads to perfect approximations in both selected realizations.

\newpage
\subsection{Heat Equation: Uncertainty Quantification}

The last numerical example presents the main advantage of \PC{} -- analytical UQ of the original model containing deterministic and random variables as proposed in section \ref{UQ}. This example is thus divided to two parts: (i) a convergence study assuming all input variables to be deterministic, which shows the efficiency of \PC{} as a surrogate model. and (ii) UQ of the given heat equation with uniformly distributed coefficient of thermal diffusivity $\mathcal{D}$ in the following form:

 \begin{align}
 \frac{\partial u(x,t)}{\partial t}&=\mathcal{D} \frac{\partial^2u(x,t)}{\partial x^2}, \qquad x \in [0, 1], \quad t \in [0, 1], \quad    \mathcal{D}\sim \pazocal{U}[0.2,0.8]\\
 u(0,t) &= u(1,t)=0, \qquad u(x,0) = \sin (\pi x) \nonumber
\end{align}

\subsubsection*{Deterministic value of coefficient of thermal diffusivity}
\begin{figure}[b!]
\centering
	\includegraphics[width=1\textwidth]{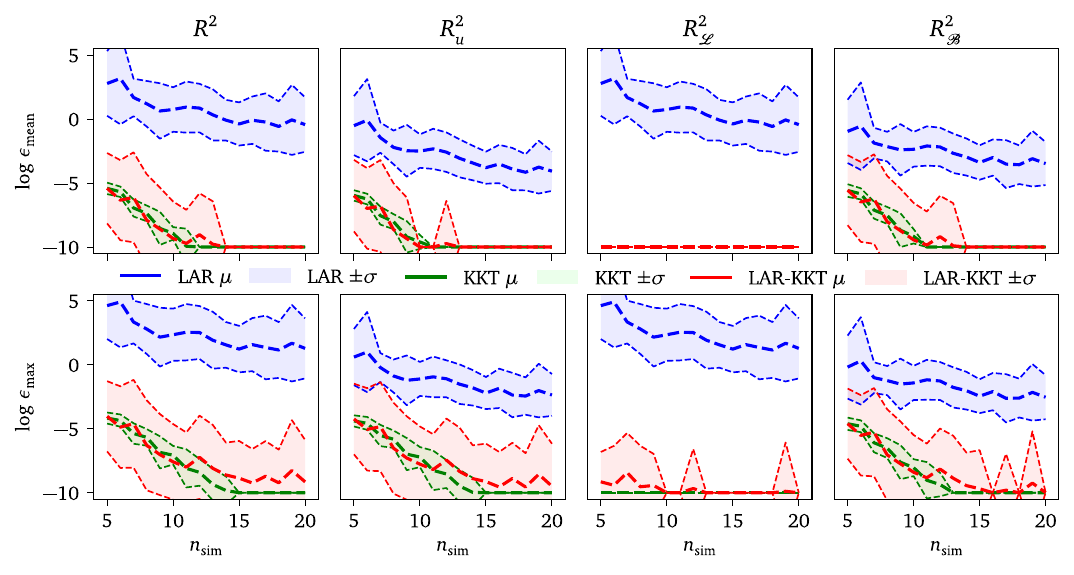}
	\caption{Obtained numerical results of heat equation for $n_{\mathrm{BC}=10}$ and increasing number of samples in the time-space domain. Top row shows mean squared errors and bottom row shows maximum squared errors obtained by simple KKT (Algorithm 1), LAR-KKT (Algorithm 2) and standard LAR method. Each column corresponds to a component of total error measure $R^2$. }
    \label{Fig. Heat Deterministic}
\end{figure}

In the first part of this example, we assume a deterministic value of the coefficient of thermal diffusivity $\mathcal{D}=0.4$. The results in Fig. \ref{Fig. Heat Deterministic} shows the convergence for $n_{\mathrm{BC}}=10$ and increasing number of samples in the space-time domain $n_{\mathrm{sim}}$. The benefit of additional information from the physical constraints can be clearly seen from the comparison of the obtained convergence plots. The convergence rate of standard LAR is significantly lower, while both  \PC{} algorithms achieve near-perfect accuracy for $n_{\mathrm{sim}}=15$. Note that LAR-KKT has higher variance since the selected sparse set of basis functions is dependent on the given ED as expected, which is clearly visible especially in the second row showing $\epsilon_{\mathrm{max}}$.

\subsubsection*{Uncertain coefficient of thermal diffusivity}

\begin{figure}[h!]
\centering
	\includegraphics[width=1\textwidth]{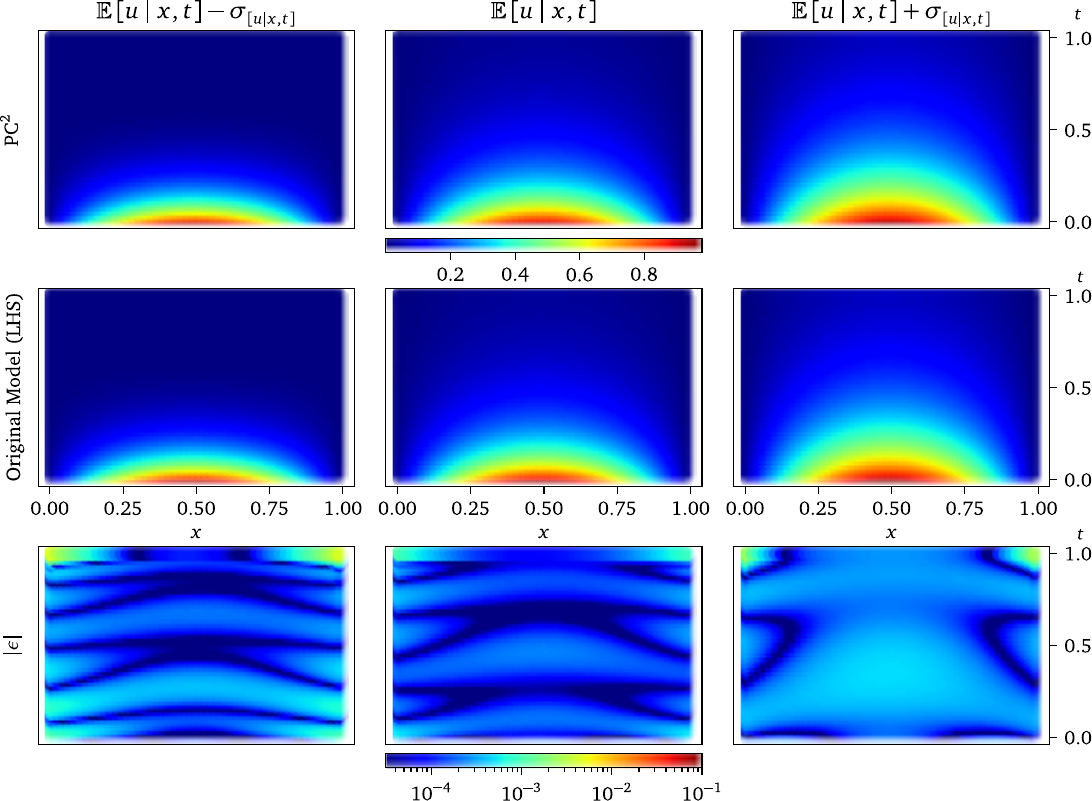}
	\caption{Estimates of the local mean values at given coordinates $\mathbb{E}\left[ u \mid x,t \right]$ (middle), and local $\pm \sigma_{\left[ u \mid x,t \right]}$ quantiles (left and right). Comparison of analytical solution by \PC{} (top), original model (middle), and their absolute difference (bottom).}
 \label{Fig. Heat Quantiles}
\end{figure}

\begin{figure}[h!]
\centering
	\includegraphics[width=1\textwidth]{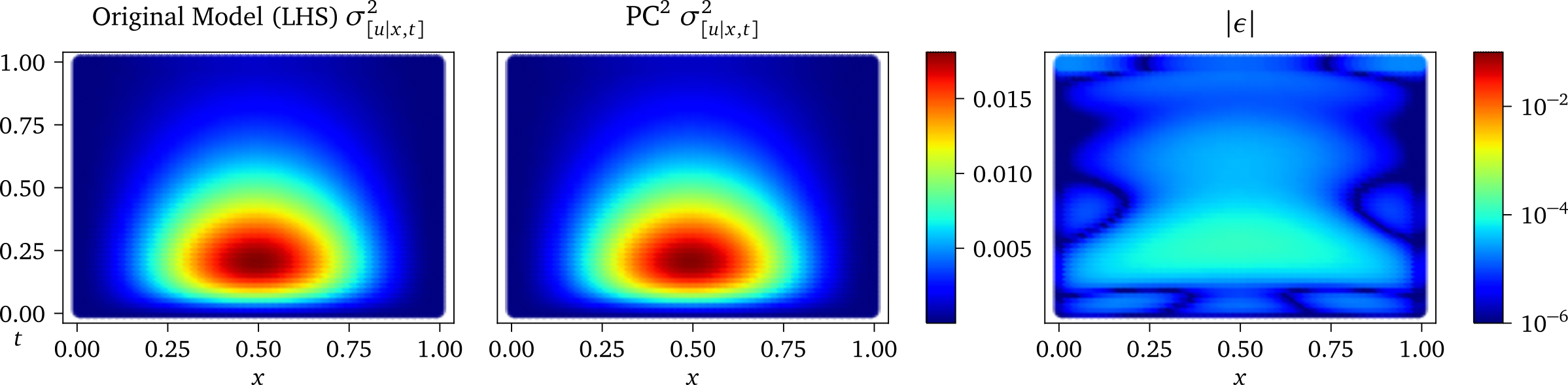}
	\caption{Estimates of the local variance  at given coordinates $\sigma^2_{\left[ u \mid x,t \right]}$. Comparison of numerical estimation by LHS (left), analytical solution by \PC{} (middle), and their absolute difference (right).}
 \label{Fig. Heat Variance}
\end{figure}

\begin{figure}[p!]
\centering
	\includegraphics[width=0.95\textwidth]{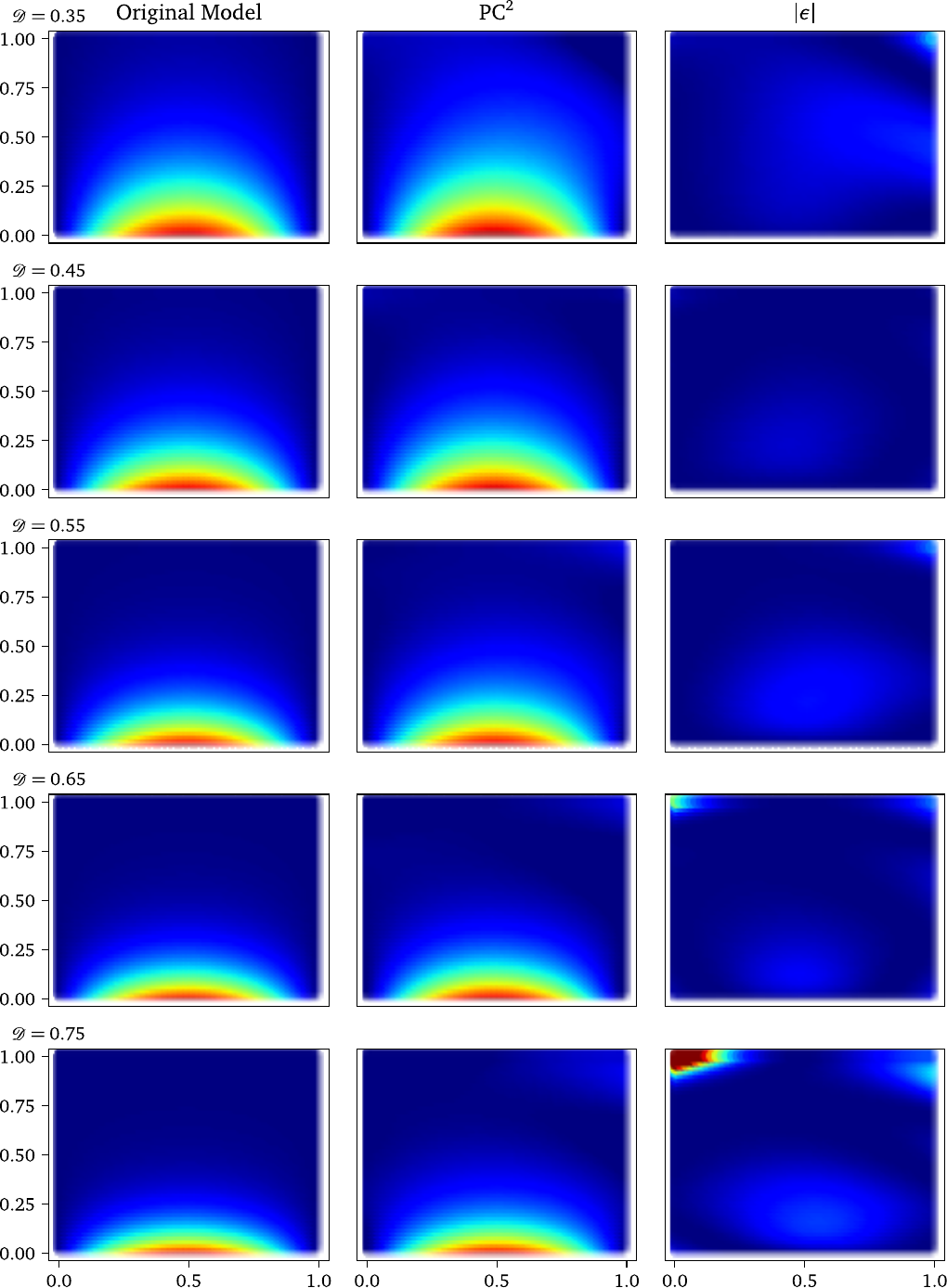}
	\caption{Approximation of heat equation by \PC{} based on $n_{\mathrm{BC}}=90$ samples. Each row shows the original function, the \PC{} approximation, and the error corresponding to a selected realization of $\mathcal{D}$.}
 \label{Fig. Heat Selected Coefficients}
\end{figure}

In the present example -- a heat equation with uncertain coefficient $\mathcal{D}\sim \pazocal{U}[0.2,0.8]$ -- there are 2 deterministic variables (space coordinate $x$ and time $t$) and one random variable $\mathcal{D}$. To efficiently quantify solution uncertainty,  \PC{} is based only on $n_{\mathrm{BC}}=90$ samples and virtual points. We do not run the deterministic simulation using samples of $\mathcal{D}$. This scenario is of practical interest, since the traditional solution of PDEs with uncertain inputs requires repetitive calculations with different realizations of the random variables which could be costly. The number of samples on the boundaries is selected relatively high to specifically present the advantages of \PC{} for UQ, since a general convergence for \PC{} was shown in the previous deterministic case. Note that the \PC{} is based only on BC and virtual samples, and thus the computational cost is not significantly affected by the assumption of uncertain $\mathcal{D}$ since it affects only the virtual point. An empty ED leads to very efficient UQ, since one can easily modify the stochastic model of the input variables -- number of random variables, their distributions etc. without re-evaluation of the ED as necessary in standard surrogate models. Once the \PC{} is available, it is possible to use it 
for UQ -- estimation of local means and quantiles (e.g. $\pm\sigma$) as depicted in Fig. \ref{Fig. Heat Quantiles}, which compares the mean solution and mean $\pm\sigma$ for \PC{} with Monte Carlo simulation using LHS with $n_{\mathrm{sim}=}10^5$ samples. Note that there are negligible errors in the local mean values. The estimated local variance can also be seen in a comparison to the LHS solution in Fig. \ref{Fig. Heat Variance}. Although the \PC{} is based only on BC and PDE, it leads to a very accurate approximation and it reflects the uncertainty of $\mathcal{D}$ very well both in predictions (see the approximations and errors for selected values of $\mathcal{D}$ in Fig. \ref{Fig. Heat Selected Coefficients}) and analytical estimation of local statistics.

\newpage

\section{Conclusions \& Further Work}
A novel methodology for construction of physics-informed non-intrusive regression-based \PCE{}, referred to as \PC{} was proposed in this paper. Given physical constraints in the form of ordinary and partial differential equations and their boundary conditions are imposed in a constrained least squares solution to assemble a KKT system with Lagrange multipliers. The proposed approach does not significantly increase the computational cost to estimate the \PCE{} coefficients, but improves accuracy considerably. The proposed solution can be further employed with existing adaptive algorithms instead of conventional ordinary least squares. An algorithm based on least angle regression is then proposed to achieve a constrained sparse solution with $p$-adaptivity. The presented approach was developed specifically for UQ of costly mathematical models of physical systems using a small ED. \PC{} allows for analytical UQ similar to standard \PCE{} by simple post-processing of the coefficients, though the \PC{} must first be reduced to exclude the influence of deterministic variables. From the obtained numerical results, it is clear that \PC{} achieves significant improvement in the accuracy of \PCE{} in terms of surrogate model prediction error, but also with respect to error in the PDE and boundary conditions for low-to-mid size ED with little additional computational cost. We compared two algorithms for \PC{} construction -- sparse solver LAR-KKT and a KKT-based \PCE{} with a full set of basis functions, and it can be concluded that while LAR-KKT does not always lead to superior results, its benefits are more significant with increasing dimension of input random vector. Note that both algorithms can be applied in tandem in practical applications, since the \PC{} using the full set of basis functions increases the computational cost of LAR-KKT just by one additional iteration. Finally, it was shown in the final numerical example that it is possible to create \PC{} using only a set of virtual samples and boundary conditions, which leads to extremely efficient UQ in comparison to other existing techniques.  

This work opens the door for several open questions that should be investigated in further research. First, the proposed approach is limited to problems with linear and weakly nonlinear differential constraints. More complicated non-linear PDEs will require advanced optimization techniques and thus \PC{} must be extended also for broader classes of applications and suitable optimization techniques should be identified. Moreover, polynomial basis functions restrict the application of \PC{} to sufficiently smooth operations, which could be alleviated by coupling the proposed method with domain-decomposition techniques \cite{novák2023active,https://doi.org/10.1002/nme.7234,WAN2005617}. Further, \PC{} will be combined with active learning techniques such as $\Theta$ criterion\cite{NOVAK2021114105} recently proposed by authors of this paper, which can be used for construction of optimal ED as well as optimal set of virtual samples.

\section*{Acknowledgments}
 The first author acknowledges the financial support provided by the Czech Science Foundation under project number 23-04712S. 
 Additionally, a part of this research was conducted during the research stay of the first author at Johns Hopkins University supported by the stipend and partnership program of Brno University of Technology for Excellence 2023. MDS and HS have been supported by the Defense Threat Reduction Agency, Award HDTRA1202000.

\bibliographystyle{elsarticle-num}
\bibliography{literatura}

\end{document}